\documentclass[sigconf,nonacm]{acmart}

\AtBeginDocument{%
  \providecommand\BibTeX{{%
    \normalfont B\kern-0.5em{\scshape i\kern-0.25em b}\kern-0.8em\TeX}}}

\setcopyright{acmlicensed}
\copyrightyear{2024}
\acmYear{2024}
\acmDOI{XXXXXXX.XXXXXXX}

\acmConference[Arxiv]{}{2024}{}
\usepackage{mdframed}
\newmdenv[leftline=false,rightline=false,backgroundcolor=gray!20,linewidth=2pt]{quotebox}

\begin{document}

\newcommand\honeyplotname{HoneyPlotNet}
\newcommand\honeyplotnameshort{HPN}
\newcommand\chartmodelname{Plot Data Model}
\newcommand\chartmodelnameshort{PDM}
\newcommand{\sharif}[1]{\textsf{\color{blue}{[{Sharif: #1}]}}}

\title{Contextual Chart Generation for Cyber Deception} 

\author{David D. Nguyen}
\email{david.nguyen@data61.csiro.au} 
\orcid{0000-0002-8243-195X}
\affiliation{
  \institution{CSIRO, Data61}
  \country{Australia}
}
\additionalaffiliation{%
    \institution{UNSW Sydney}
    }
\additionalaffiliation{%
    \institution{Cybersecurity CRC}
    }
    
\author{David Liebowitz}
\email{david.liebowitz@penten.com} 
\affiliation{
  \institution{Penten}
  \country{Australia}
}
\authornotemark[1]

\author{Surya Nepal}
\email{surya.nepal@data61.csiro.au} 
\orcid{0000-0002-3289-6599}
\affiliation{
  \institution{CSIRO, Data61}
  \country{Australia}
}
\authornotemark[2]

\author{Salil S. Kanhere}
\email{salil.kanhere@unsw.edu.au} 
\orcid{0000-0002-1835-3475}
\affiliation{
  \institution{UNSW Sydney}
  \country{Australia}
}
\authornotemark[2]

\author{Sharif Abuadbba}
\email{sharif.abuadbba@data61.csiro.au} 
\orcid{0000-0001-9695-7947}
\affiliation{
  \institution{CSIRO, Data61}
  \country{Australia}
}
\authornotemark[2]

\renewcommand{\shortauthors}{Nguyen et al.}

\begin{abstract}
Honeyfiles are security assets designed to attract and detect intruders on compromised systems.
Honeyfiles are a type of honeypot that  mimic real, sensitive documents, creating the illusion of the presence of valuable data. 
Interaction with a honeyfile reveals the presence of an intruder, and can provide insights into their goals and intentions. 
Their practical use, however, is limited by the time, cost and effort associated with manually creating realistic content.
The introduction of large language models has made high-quality text generation accessible, but honeyfiles contain a variety of content including charts, tables and images.
This content needs to be plausible and realistic, as well as semantically consistent both within honeyfiles and with the real documents they mimic, to successfully deceive an intruder.

In this paper, we focus on an important component of the honeyfile content generation problem: document charts.
Charts are ubiquitous in corporate documents and are commonly used to communicate quantitative and scientific data.
Existing image generation models, such as DALL-E, are rather prone to generating charts with incomprehensible text and unconvincing data. 
We take a multi-modal approach to this problem by combining two purpose-built generative models: a \textit{multitask Transformer} and a \textit{specialized multi-head autoencoder}.
The Transformer generates realistic captions and plot text, while the autoencoder generates the underlying tabular data for the plot.
We refer to this new model as \honeyplotname.

To advance the field of automated honeyplot generation, we also release a new document-chart dataset and propose a novel metric Keyword Semantic Matching (KSM).
This metric measures the semantic consistency between keywords of a corpus and a smaller bag of words.
Extensive experiments demonstrate excellent performance against multiple large language models, including ChatGPT and GPT4.
Our code has been anonymously released on GitFront \href{https://gitfront.io/r/user-5701462/FDq8HjWdDaUz/honeyplotnet/}{[Link]}.

\end{abstract}

\maketitle

\section{Introduction}
\label{sec:intro}

Despite the significant investment in intrusion detection systems over the past decade, the average time to identify and contain a security breach is 323 days, and its average cost is \$4.35 million, according to the IBM's \emph{Cost of a Data Breach Report 2022} \cite{ibm2022cost}.
This growing problem highlights the need for intrusion detection techniques that can identify unauthorised interaction within compromised systems. 
\textbf{Honeypots} are one of the most powerful cybersecurity tools that can be employed to address this challenge~\cite{spitzner2003-1-honeypots}. 
Designed to mimic real devices or resources, honeypots are of no use to legitimate users and thus more likely to attract the attention of intruders reconnoitring a system. They can be monitored for access or other forms of interaction as a reliable indicator of a breach. 
More sophisticated honeypots allow for greater engagement, potentially yielding intelligence regarding an intruder's intent, tools or malicious payloads~\cite{kelly2021comparative}. 
Many organizations conduct research into and commercialize honeypots and related cyber deception technology, including Attivo, Countercraft, Thinkst and Penten. 

Fake documents or \emph{honeyfiles}~\cite{yuill2004honeyfiles} are a particularly useful type of honeypot: documents are ubiquitous and often contain valuable information such as intellectual property and financial data. 
Honeyfiles are easy to deploy and can be crafted to contain topics from sensitive documents or content that matches the interests of suspected threats\cite{salem2011,voris2013bait}. 
Honeyfiles placed in a document repository or file system can be monitored for access or exfiltration as a breach detection 
mechanism. 
The choice of topic or search terms used by an adversary to look for documents in a repository can also provide insight into their intent and interests~\cite{timmer2022tsm}. 
The key to successful honeyfile use is \emph{realism}, in the sense that the appearance and content of a honeyfile accurately mimic real documents. 

Ideally, honeyfiles are generated automatically so that they can be created in abundant variety with minimal cost and effort~\cite{liebowitz2021deception}. 
The development of GPT~\cite{radford2018improving} and other Transformer-family language models has made text generation accessible, but text content is not the only aspect of a honeyfile that must be realistic. 
Many technical and financial documents include quantitative data presented in plots or charts.
The presence of charts in honeyfiles is thus a key component of realism, not only because they are part of the mimicry of actual documents, but also because they are visually salient. 
The overall appearance of a document is critical to the perception of realism, particularly in the exploratory phase of intrusion when an attacker is searching for data~\cite{bowen2009baiting,ben2011decoy,liebowitz2021deception}. 
This might involve directory browsing or keyword searches in a document repository or knowledge base and rapid scrutiny of document contents or page snapshots returned by many search engines. 
This \emph{preview realism}~\cite{moore2022modelling} must persuade an intruder that a document is real to elicit interaction with the honeyfile. 
Further engagement, and potential download or exfiltration, can result in closer scrutiny. 
At this later stage, realism should extend to plausible text and charts that are semantically consistent with the synthesised text. 

\begin{figure*}[h]
    \centering
    \includegraphics[width=\textwidth, scale=0.1]{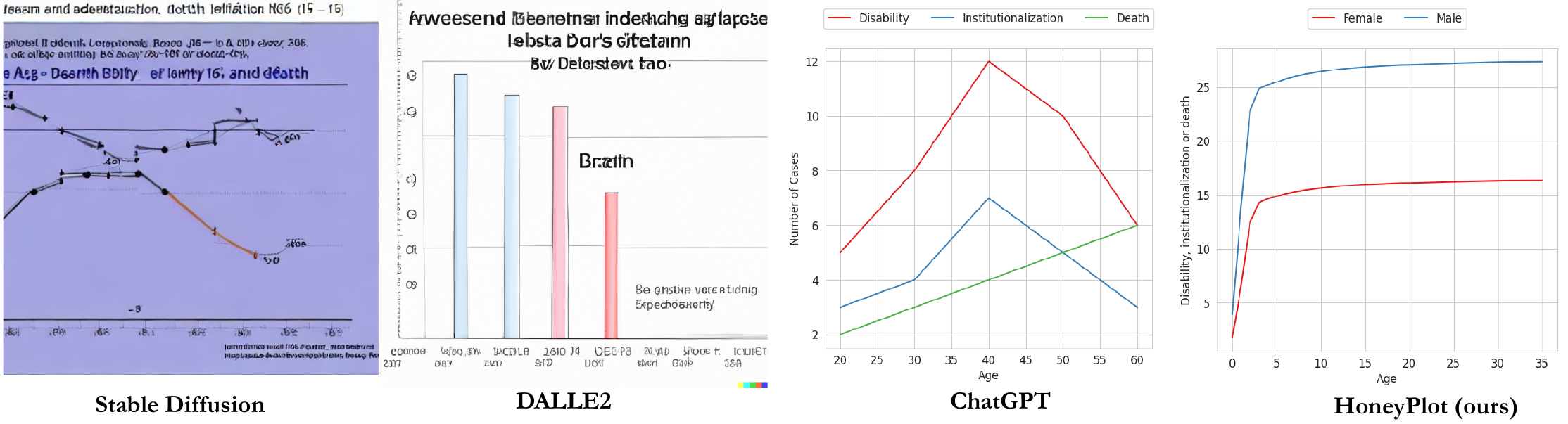}
    \caption{Honeyplots generated by different image and language models using the same prompt/caption: "Age and gender difference of disability, institutionalization and death (n = 1560)".} 
    \label{fig:pixelcomparison}
\end{figure*}

The major challenge of adding charts to honeyfiles is balancing the required level of semantic consistency and labour cost.
Hand-crafting semantically consistent charts for each region of document text is not scalable and is very expensive labour-wise \cite{bowen2009baiting}.
At the opposite end of the spectrum, randomly generating charts is cheap, but is likely to result in charts inconsistent with the document topic and content which raise the suspicion of an intruder.
A promising approach that can improve scalability and realism is to employ \emph{generative models}, a class of models that learn the underlying distribution of data which allows it to generate new, unseen data. 
However, our investigation finds that existing large language~ \cite{radford2015unsupervised,gpt4techreport} or image generative models~  \cite{esser2021taming,villegas2022phenaki} are unable to create convincing charts for the following reasons:
\begin{enumerate}
    \item \textbf{Incomprehensible and nonsensical text.} Image generative models, while adept at creating visuals, lack an understanding of the semantic meaning of the text they produce (see Fig~\ref{fig:pixelcomparison}). This limitation often leads to incomprehensible or irrelevant text, increasing the risk of detection by attackers.  
    \item \textbf{Unconvincing visuals.} Image generative models, although capable of generating charts, often produce visually lacking authenticity. The style, content, and data these charts represent typically appear artificial, making them easily identifiable as fake in documents. This can also be seen in Fig~\ref{fig:pixelcomparison}).
    \item \textbf{Short attention spans.} Text-to-image models, trained primarily on brief captions \cite{ramesh2021zero}, tend to overlook extensive text sections in longer documents due to their shorter attention spans.
    \item \textbf{Text Only.} Large language models (LLMs), such as ChatGPT \cite{gpt4techreport}, are unable to produce anything except text without the assistance of an external tool.  
\end{enumerate}

The absence of document-to-chart generation models in the current literature can be attributed to two core issues. 
First, there is a lack of open-source datasets specifically designed for document chart generation. 
While there are multiple datasets that contain documents or charts, none combine the two.
Second, there is no widely accepted framework for evaluating the realism of generated charts. 
Complexity arises from the need to accurately represent both textual and numerical data in charts, which cannot be easily assessed by a single metric.

This paper answers the following questions to address these challenges and the limitations of existing generative models:
\begin{enumerate}
    \item Can we develop a generative model capable of synthesizing realistic charts based on the long document text?
    \item Is it feasible to construct a dataset that pairs documents with charts, aiding the training of this generative model?
    \item What metrics are needed to measure the quality of charts found in documents?
\end{enumerate}
Addressing these questions will tackle major hurdles, such as scalability and semantic consistency, enhancing the quality of existing cyber deception systems. 

In this paper, we introduce the \honeyplotname, a novel combination of generative models for document-to-chart generation.
The architecture combines a large language model (LLM) and a multi-head vector quantization variational autoencoder (VQVAE), as illustrated in Fig~\ref{fig:archoverview}.
The LLM is a multimodal multitask Transformer, which generates textual content based on the surrounding document text. 
The multi-head VQVAE can synthesize the underlying data for multiple chart types which is jointly trained end-to-end.
The outputs of both generative models are then fed into chart visualization software for rendering. 
This process ensures that the honeyplots have no rendering defects that could expose their artificial nature.
The resulting honeyplots are visually accurate representations of the generated data and blend into the surrounding honeyfile.

The benefit of our approach is that all honeyplot content is generated using a \textit{single unified architecture}.
Developing a unified architecture is challenging because it requires careful alignment between the representational spaces of each modality to create semantic consistency between the captions, plot text and data.
Furthermore, it necessitates a specialized training strategy and architecture design to accommodate a large number of generation tasks, such as captions and axis titles and different types of charts like bars and lines.
Employing a single unified architecture brings notable benefits, including improved performance, reduced training times and streamlined deployment in real-world scenarios.

We also introduce the first chart generation dataset which is a valuable resource for the cyber deception community.
The dataset combines chart data released by other researchers \cite{davila2021icpr} and text harvested from PubMedCentral OpenAccess.
This dataset is used to benchmark several variants of our model against other large language models, including ChatGPT \cite{brown2020language} and GPT4 \cite{gpt4techreport}.

Furthermore, we propose a novel metric called Keyword Semantic Matching which fills a critical gap in honeyfile evaluation.
The Keyword Semantic Matching (KSM) metric measures the semantic similarity between the generated chart text and the local document text.
This is used in conjunction with ROUGE \cite{lin2004rouge}, a widely used summarization metric that measures overlapping n-grams. 
Our results confirm that \honeyplotname \ results in high-quality captions and charts with high semantic similarity to the document text.

Our contributions are summarized as follows:
\begin{enumerate}
    \item \textbf{\honeyplotname.} We introduce a generative model that synthesizes realistic charts based on long document text.
    This architecture unifies a large language model and a specialized multi-head autoencoder to reduce training computational requirements and deployment complexity.
    We propose a training strategy that carefully aligns the language and chart data representation space.
    To the extent of our knowledge, this is also the first model proposed for honeyplot generation which addresses the four identified limitations above.
    \item \textbf{Document-Chart Dataset.} We open-source the first dataset containing 5418  document-chart pairs. This dataset contains 5 different chart types: line, scatter, bar charts and box plots. This release will progress the field by helping others develop generative models for cyber deception. 
    \item \textbf{Keyword Semantic Matching.} We propose a new metric to evaluate the semantic consistency between a document corpus (a large body of text) and chart text (a smaller collection of words). This metric fills a gap in current NLP benchmarks by effectively assessing the semantic quality of text used in charts. 
\end{enumerate}

This paper begins with a review of related literature in Section~\ref{plots:sec:related}. 
We formulate our problem and introduce our chart generation architecture in Section~\ref{plots:sec:chartgen}.
We outline our evaluation framework in Section~\ref{plots:sec:eval} and present our experimental results in Section~\ref{plots:sec:results}.
We conclude our findings and provide directions for future work in Section~\ref{plots:sec:conclusion}.


\section{Related Work}
\label{plots:sec:related}
This section explores recent advancements in tabular models and Transformers for multitask and multimodal learning, focusing on their application and limitations in synthesizing honeyplots from documents.
\subsection{Tabular Models}
The underlying plot data of charts can be structured in a tabular format. 
Tabular data generation is an active area of research  \cite{kotelnikov2022tabddpm, xu2019modeling,zhang2021ganblr,zhao2021ctab,jordon2018pate,grinsztajn2022tree}  with various approaches proposed over the years.
Gradient Boosting Decision Trees (GBDT) have been the state-of-the-art method in the field for the past decade, leading with algorithms like XGBoost \cite{chen2016xgboost}, CatBoost \cite{prokhorenkova2018catboost} and LightGBM \cite{ke2017lightgbm}.
More recent works have suggested that deep learning architectures, such as TabTransformer \cite{huang2020tabtransformer}, can achieve comparable or superior performance than GBDT. 
Despite these improvements, we argue that these algorithms are not well-suited for honeypot generation due to two fundamental assumptions made by tabular models. Below, we list each problem and contrast what tabular models provide vs what is needed to create charts for documents.  

\paragraph{\textbf{Problem 1: Tabular Models are restricted to only one data structure.}}
    \begin{itemize}
        \item \textbf{Tabular Models.} The fundamental design of existing tabular models \textbf{assumes} that all entries of the dataset have the same data structure \cite{kotelnikov2022tabddpm,xu2019modeling,zhang2021ganblr,zhao2021ctab,jordon2018pate,huang2020tabtransformer}. 
        Each entry is assumed to have the same number of variables, variable types and distributions for each column. 
        \item \textbf{Honeyplot Models.} The underlying data structure \textbf{of each chart} can be effectively regarded as its own tabular dataset. 
        For example, a scientific paper containing $N$ plots would use $N$ different underlying datasets because each plot would illustrate a different topic. 
        The underlying dataset of one plot may represent blood haemoglobin percentages vs allele type, while the next may represent vaccination rates across each country. 
        
        For this reason, we \textbf{cannot assume} that the underlying data of all plots in the dataset will contain the \textbf{same data structure}.
        Each plot in the dataset will vary in the number of variables, variable types and distributions across each column.
        Consequently, a different tabular model must be trained for each plot, which creates various downstream problems.
        To address this, our \emph{Plot Data Model} produces different dataset structures for each plot.
    \end{itemize}

\paragraph{\textbf{Problem 2: Tabular models cannot consider the local context for categorical data}}
    \begin{itemize}
        \item \textbf{Tabular Models.} In standard tabular models \cite{kotelnikov2022tabddpm,xu2019modeling,zhang2021ganblr,zhao2021ctab,jordon2018pate,huang2020tabtransformer}, categorical data is treated simplistically. Each unique category is assigned a distinct index. For instance, in a column with animal types, "Dog" might be assigned index "2" in a list ["Cat", "Cow", "Dog"].
        This approach ignores the context or meaning behind these categories. It treats them as mere labels without considering any semantic relationship or relevance to the document’s content.
        \item \textbf{Honeyplots.} In real-world documents, the categories in charts often relate directly to the document's topic or the text surrounding the chart.
        Our model uses pre-trained language embeddings to generate categorical data based on the surrounding text to address this.
         This method ensures that the categories generated for a honeyplot are contextually aligned with the document's content. 
         For example, if the document talks about environmental issues, the honeyplot's categories will be relevant to environmental themes rather than random or unrelated topics.
        By making the categorical data contextually relevant, the honeyplot becomes more convincing and harder for attackers to identify as a decoy.
    \end{itemize}

To summarize, the assumptions made by traditional tabular models regarding dataset structure and categorical data are not suitable for honeyplot generation. 
Addressing these limitations requires a specialized approach which we will discuss next.

\subsection{Transformers for multitask and multimodal learning}
The Transformer, introduced by Vaswani et al.  \cite{vaswani2017attention}, is a neural network architecture first proposed for solving sequence-to-sequence language modelling problems.
The seminal paper demonstrated the scalability and effectiveness of the multi-head attention mechanism in capturing word relationships within sentence and led to state-of-the-art performance on machine translation tasks.
Subsequent models, such as GPT \cite{radford2018improving,radford2019language,gpt4techreport}, BERT \cite{devlin2018bert}, XLNet \cite{yang2019xlnet}, RoBERTa \cite{liu2019roberta}, and T5 \cite{raffel2020exploring} have further demonstrated the strength and scalability of Transformers.
By pretraining these models on vast corpora, they acquire robust language representations that can be effectively fine-tuned for a wide range of downstream tasks.

These extensions of the Transformer architecture fine-tune a specific set of parameters for each of the different downstream tasks.
For instance, in the case of BERT, multiple pre-trained models are fine-tuned individually for different language tasks. 
Similarly, the more recent T5 model fine-tunes distinct sets of parameters for each specific task. 
In contrast, we adopt a multitask learning approach \cite{caruana1997multitask,crawshaw2020multi,sogaard2016deep,mahabadi2021parameter} by fine-tuning all chart language tasks simultaneously. 
This approach not only saves training time but also reduces the overall number of parameters, resulting in cheaper and faster real-world deployment.

Transformers have also been applied towards many multi-modal problems in the vision and language domain.
These multi-modal problems can be categorized by the domain types of the input and output.
For instance, visual question-answering tasks require processing both image and language modalities as input and produce text outputs \cite{lu2019vilbert,li2019visualbert,tan2019lxmert,gpt4techreport}.
Conversely, text-to-vision synthesis problems take language as input and produce visual outputs, such as images or videos \cite{nichol2021glide,villegas2022phenaki,saharia2022photorealistic}. 
In the case of our honeyplot problem, it involves taking language as input and producing both language and chart data as output.

Previous research in multimodal Transformers often utilized specific architectures tailored to individual modalities. 
However, recent studies have demonstrated the feasibility of employing a unified Transformer architecture that can handle all modalities  \cite{kaiser2017one,hu2021unit,lu202012,gpt4techreport}. 
In line with this design approach, our work adopts a shared language encoder while employing separate decoders for each output modality. 
Our work is also the first multimodal Transformer demonstrated to address document-chart generation.

\section{\honeyplotname \ System Design}
\label{plots:sec:chartgen}
In this section, we provide a detailed understanding of the design and architecture of our novel generative model, HoneyPlotNet.
The design goals were to achieve the following characteristics:
\begin{itemize}
    \item Semantically consistent text across the honeyplot and document.
    \item Realistic tabular data that relates to local text.
    \item Convincing chart visualization without any defects. 
\end{itemize}
which are not all present in any existing model.
To achieve these goals, our approach is to unify generative models for both language and non-language modalities.
We describe this architecture next. 

\begin{figure*}[t]
  \centering
  \includegraphics[width=\linewidth]{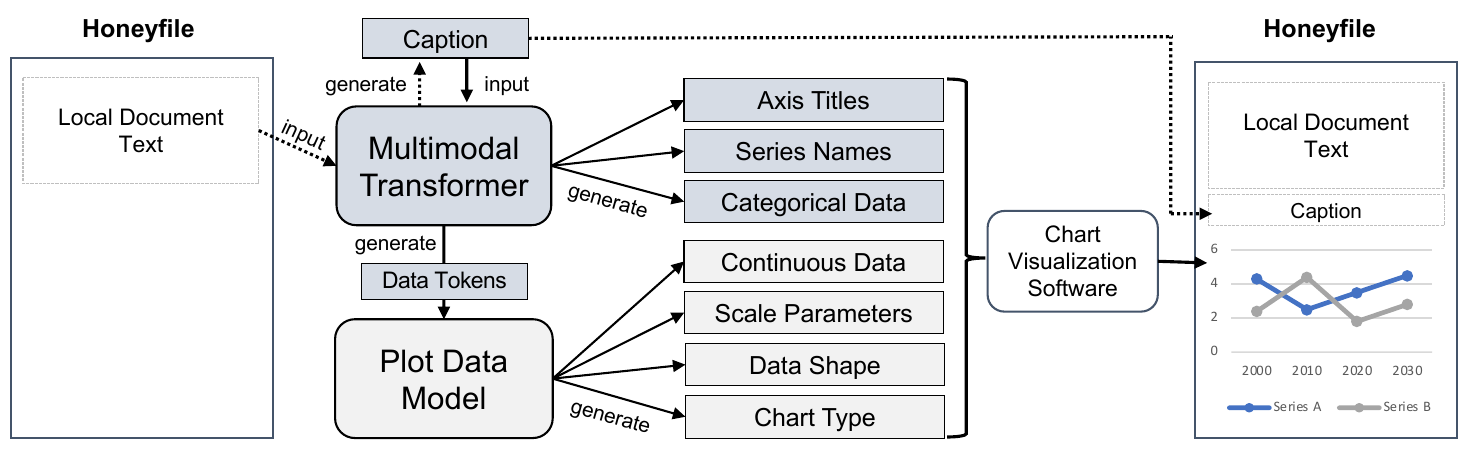}
  \caption{Architecture overview of the \honeyplotname. The document text is fed into a Transformer language model to generate captions (dotted lines). The captions are fed back to generate chart text and data tokens (solid lines). Tokens are passed into the Plot Data Model, which generates the continuous chart data. }
   \label{fig:archoverview}
\end{figure*}




\subsection{Architecture Overview}
\label{sec:archoverview}

Fig~\ref{fig:archoverview} illustrates our proposed chart generation architecture.
This approach uniquely connects the document text, caption and chart data to create contextual consistency.
The architecture is composed of two core models:
\begin{enumerate}
    \item \textbf{\chartmodelname} \ This model is responsible for generating the chart type and continuous chart data. 
    The model is implemented using a multi-head hierarchical vector quantization autoencoder that employs a latent codebook of $K$ vectors.
    In this approach, each chart data instance of the dataset is mapped to a discrete latent variable, which refers to the codebook indices.
    We refer to these as the \emph{data tokens} to distinguish them from language tokens, typically used by language models.
    \item \textbf{Multimodal Transformer Model}. This model is responsible for taking the document text as input and outputting five different chart attributes: captions, categorical values, series names, axis names and chart data tokens.
    The base design is a sequence-to-sequence Transformer with a shared encoder and two decoders for each modality: natural language and data. 
    This model is trained using multi-task learning to reduce computational costs and improve performance.
\end{enumerate}
We further describe both models in the following sections.

\subsection{\chartmodelname}
\label{sec:dae}
This section describes a model that learns to generate the chart type and continuous chart data. 
This model addresses two major challenges encountered when dealing with chart data.

The first challenge is that the scale of raw chart data can vary wildly, with values ranging from percentages to millions.
To address this, we propose a unique pre-processing step in \textbf{Section~\ref{sec:preprocess}} that involves splitting the raw chart data into the normalized data and scale parameters. 
The second challenge is that the input data structure for different chart types varies in shape and dimensionality.

\textit{The key novelty of our approach is that all chart types and a wide variety of data structures are learned using a single model.}
The base architecture is derived from the hierarchical Vector Quantization Variational Autoencoder (VQVAE) \cite{van2017neural,dhariwal2020jukebox} which is briefly discussed in \textbf{Section~\ref{sec:vq}}.
To learn multiple chart type formats and data components simultaneously, we propose a novel multi-head encoder (\textbf{Section~\ref{sec:encoder}}) and decoder (\textbf{Section~\ref{sec:decoder}}).


\subsubsection{\textbf{Pre-processing}}
\label{sec:preprocess}
Next, we employ data processing to address the unique scaling issue found in charts. 
Data that appears in charts can occur at nearly any scale: probabilities might be charted as fractions, and sales numbers or astronomical distances could be in the millions. 
To train a model to represent and generate data at such widely varying scales, the model should use \emph{normalised} data.
We also need to represent and generate the \emph{scale parameters} that can map between normalised and chart scales.

The data in charts can also have varying sizes: one may display one line with 15 points, and another may display ten lines with 30 points.
To manage this, the decoder learns to predict the overall data shape.
Below, we discuss how to construct each of these data components from the raw unnormalized chart data.

\paragraph{\textbf{Raw Data Structure}, $\mathbf{x} \in \mathbb{R}^{H \times W \times D}$}
Here, we describe the tensor representation of raw chart data, which is the starting point of our preprocessing process.
The raw continuous data of graphical charts can be generalized as a 3-tensor with at most $H$ rows, $W$ columns and $D$ features.
Each instance may have a different number of columns and rows.
Using line charts as an example, the number of columns reflects the number of points in a single line, and the number of rows reflects the number of lines.

Each chart type uses a different number of features.
The bar chart contains one feature representing the height of the vertical bar (or width of a horizontal bar).
Line and scatter charts contain two features, representing the (x,y) coordinates of a single point.
Box charts contain five features: minimum, first quartile, median, third quartile and maximum.

\paragraph{\textbf{Data Normalization}, $\mathbf{x}^\prime \in \mathbb{R}^{H \times W \times D}$}
Next we describe how to normalize raw chart data for each chart type. 
 Chart data is normalized by scaling the values to the range [0,1] using the minimum $\textbf{x}_{\min}$ and maximum $\textbf{x}_{\max}$ values over each group of rows and features.

Box charts are processed differently from other charts.
The normalized data for box charts contains 5 features: the \emph{normalized minimum} and the \emph{absolute difference between the normalized minimum} and the normalized value for each of the other four features: first quartile, median, third quartile and maximum.
This ensures that every unit of $\mathbf{x}^\prime$ is between 0 and 1.

\paragraph{\textbf{Parameter Scaling}, $\mathbf{t} \in \mathbb{R}^{H \times 2D}$}
 We compute two parameters for each row and feature: a minimum value $\mathbf{t}_{\min}$ and range  $\mathbf{t}_{range}$.
To calculate these parameters, we scale $\textbf{x}_{\min}$ and $\textbf{x}_{\max}$ between $(\gamma, +\infty)$ using the common logarithm, $\log_{10}$:
\begin{equation}
\begin{aligned}
    \mathbf{t}_{\min} &= \log_{10}(\textbf{x}_{\min} + (1.1 + \gamma))\\
    \mathbf{t}_{\max} &= \log_{10}(\textbf{x}_{\max} + (1.1 + \gamma)) \\ 
    \mathbf{t}_{range} &= \mathbf{t}_{\max} - \mathbf{t}_{\min}
    \label{eq:norm}
\end{aligned}
\end{equation}
where $\gamma$ also acts as a floor and is typically a very small number close to zero, such as $\gamma=1e{-4}$.
There are only two parameters in total for box charts: the minimum and the range for the normalized minimum.

\subsubsection{\textbf{VQVAE}}
\label{sec:vq} 
Our model is based on the VQVAE that employs a discrete latent space.
This characteristic allows seamless integration with existing language language models that utilize latent embeddings.
The VQVAE is a lossy compression algorithm that encodes chart data to the nearest vectors in a latent codebook of $K$ vectors, $\mathbf{e}_k \in \mathbb{R}^d$.
During training, the distance between encoded chart data and the nearest codebook vectors is minimized, resulting in clusters of similarity in latent space.
As the number of codebook vectors is limited, this algorithm forces the model to find patterns in the data. 

\begin{figure*}[t]
  \centering
  \includegraphics[width=\linewidth]{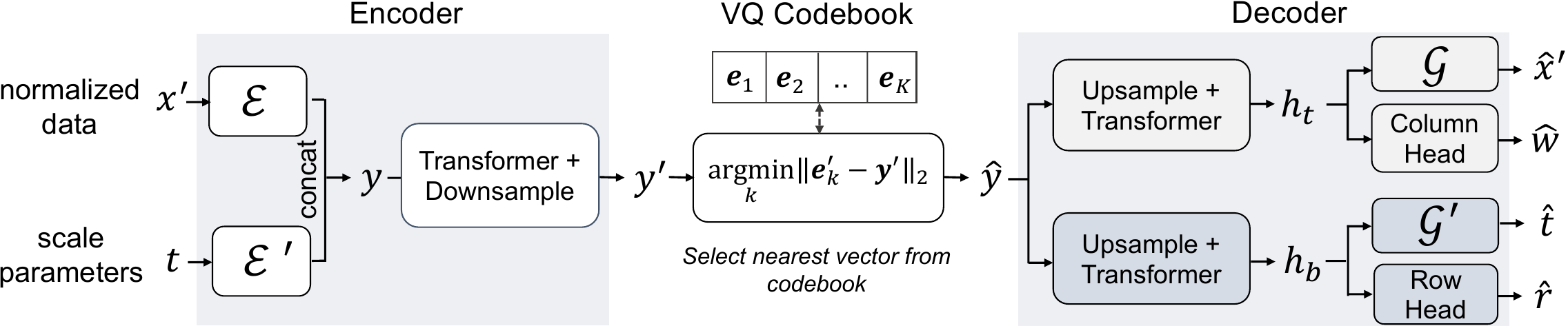}
  \caption{Overview of the Plot Data Model, which combines the VQ framework with multi-head encoder and decoder. This model is responsible for generating continuous data values for multiple chart types. See Section~\ref{sec:encoder} for the encoder and Section~\ref{sec:decoder} for the decoder.}
   \label{fig:vae}
\end{figure*} 

As shown in Fig~\ref{fig:vae}, a pass through the encoder results in each encoded vector being replaced by the nearest vector in the codebook.
The sequence of quantized vectors $\mathbf{\hat{y}}$ is fed to the decoder, which learns to reconstruct the original data.
The decoder is trained by minimizing the mean square error between the original data and its reconstruction.
We design purpose-built encoders and decoders to handle chart data, as we will discuss next. 

\subsubsection{\textbf{Multi-Head Encoder}}
\label{sec:encoder}
Our proposed encoder ensures that multiple chart types can be simultaneously trained under the same framework 
As seen in Fig~\ref{fig:vae}, there are two separate pathways for normalized data and scale parameters.

\paragraph{Top path} The normalized data is passed into a multi-head group of $D$ chart-specific heads, $\mathcal{E}=(\mathcal{E}_1,\ldots,\mathcal{E}_D)$.
Recall that different chart types have different features for their normalized data.
Each head is a linear neural network layer that projects the normalised data's dimensionality from a range of $D$ features to an equivalent dimensionality $d$.

\paragraph{Bottom path} The scale parameters are passed into another multi-head group of $D$ chart-specific heads, $\mathcal{E}^\prime=(\mathcal{E}^\prime_1,\ldots,\mathcal{E}^\prime_n)$.
Each head also only encodes the scale parameters based on the number of features and reduces their dimensionality to $d$. 

\paragraph{Fusion} To encode inductive bias, learnable positional embeddings are added to the outputs of each head.
These vectors are then concatenated and flattened along the row and column dimension into a sequence of vectors $\mathbf{y} \in \mathbb{R}^{S \times d}$.

In the final encoding step, $\mathbf{y}$ is down-sampled using a Transformer encoder layer and a down-sampling network that outputs a sequence of encoded vectors, $\mathbf{y}^{\prime} \in \mathbb{R}^{s \times d}$.
This sequence is passed to the latent codebook bottleneck for quantization.

\subsubsection{\textbf{Multi-Head Conditional Decoder}}
\label{sec:decoder}
The decoder can be thought of as an encoder in reverse, with several key differences. 
First, the decoder is trained conditionally by appending a chart-type embedding to the quantized vectors.
Second, the decoder must predict the data shape: the true number of rows $h$ and columns $w$ of each instance. 
As seen in Fig~\ref{fig:vae}, the decoder contains two separate pathways that are fed the same quantized vectors.

\paragraph{Top path}
This pathway is responsible for reconstructing the normalized data and predicting the number of columns.
The pathway begins with an up-sampling 1-D convolutional neural network layer and Transformer, which autoregressively predicts elements of a hidden matrix $\mathbf{h}_t \in \mathbb{R}^{H \times W \times d}$ in a raster scan pattern.

 At each autoregressive step, the hidden matrix is fed to a multi-head group, $\mathcal{G}=(\mathcal{G}_1,\ldots,\mathcal{G}_D)$ which outputs a reconstruction of the normalized data $\mathbf{\hat{x}^\prime}$.
Each chart-specific head uses a sigmoid activation function to ensure values are in the $[0,1]$ range.
The heads are trained by minimizing the mean squared error between the reconstructed and original normalized data.

To predict the number of columns, the same hidden matrix is passed to the \emph{column head}, which outputs a vector $\mathbf{\hat{w}} \in \mathbb{R}^{W}$.
From left to right, each element of this vector represents the probability of adding another column to the final data shape.
The column head is a single layer with a sigmoid activation trained with binary cross-entropy loss.

\paragraph{Bottom path}
This pathway is responsible for reconstructing the scale parameters and predicting the number of rows. 
This pathway also begins with an up-sampling layer and Transformer, which autoregressively predicts a hidden matrix $\mathbf{h}_b \in \mathbb{R}^{H \times d}$ sequentially across the $H$ dimension.

At each autoregressive step, the scale parameters are reconstructed by passing the hidden matrix to the multi-head group $\mathcal{G}^\prime=(\mathcal{G}^\prime_1,\ldots,\mathcal{G}^\prime_D)$ with ReLU activation layers.
We train the multi-head group by minimizing smooth L1 loss \cite{girshick2015fast} between the reconstructed $\mathbf{\hat{t}} \in \mathbb{R}^{H \times 2D}$ and original scale parameters.
Due to the large presence of outliers, we find that this loss function is more stable than mean squared error.

To predict the number of rows,  the same hidden matrix is passed to the \emph{row head}, which outputs a vector $\mathbf{\hat{r}} \in \mathbb{R}^{H}$.
The row head is implemented as a single layer with a sigmoid activation and trained with binary cross-entropy loss. 

\subsection{Multimodal Multitask Transformer Model}
\label{sec:charttextmodel}
Here, we describe the multimodal architecture and multitask learning framework which utilizes a pre-trained large Transformer language model.
This model learns five different tasks across two modalities: 
\begin{enumerate}
    \item Generating \emph{captions} based on the document text.
    \item Generating \emph{categorical data}, \emph{series names} and \emph{axis titles} based on the caption.
    \item Generating \emph{data tokens} based on the captions.
\end{enumerate}
\textit{The novelty of our multimodal and multitask approach is that all tasks are learned using a single model.}
Building a single model for multiple modalities is challenging because each modality has its own unique representations.
Constructing meaningful alignments between language and chart data is nontrivial and requires a specialized training strategy and unique architecture, as discussed in the next subsection.
Furthermore, using a single model significantly reduces training times and simplifies deployment in the real world. 

\subsubsection{\textbf{Architecture.}}

The base architecture design is a sequence-to-sequence Transformer with a shared encoder and two decoders for each modality: natural language and chart data. 
Each decoder accepts the encoder's last hidden state and is fitted with separate linear heads.
Each head predicts the next language or data token and is trained using cross-entropy loss. 
Separate decoders are employed because training stability improves when different modalities utilize different embedding matrices. 
The shared encoder and language decoder are initialized with pre-trained weights; however, the data decoder is randomly initialized.

\begin{figure}[h]
  \centering
  \includegraphics[width=8cm]{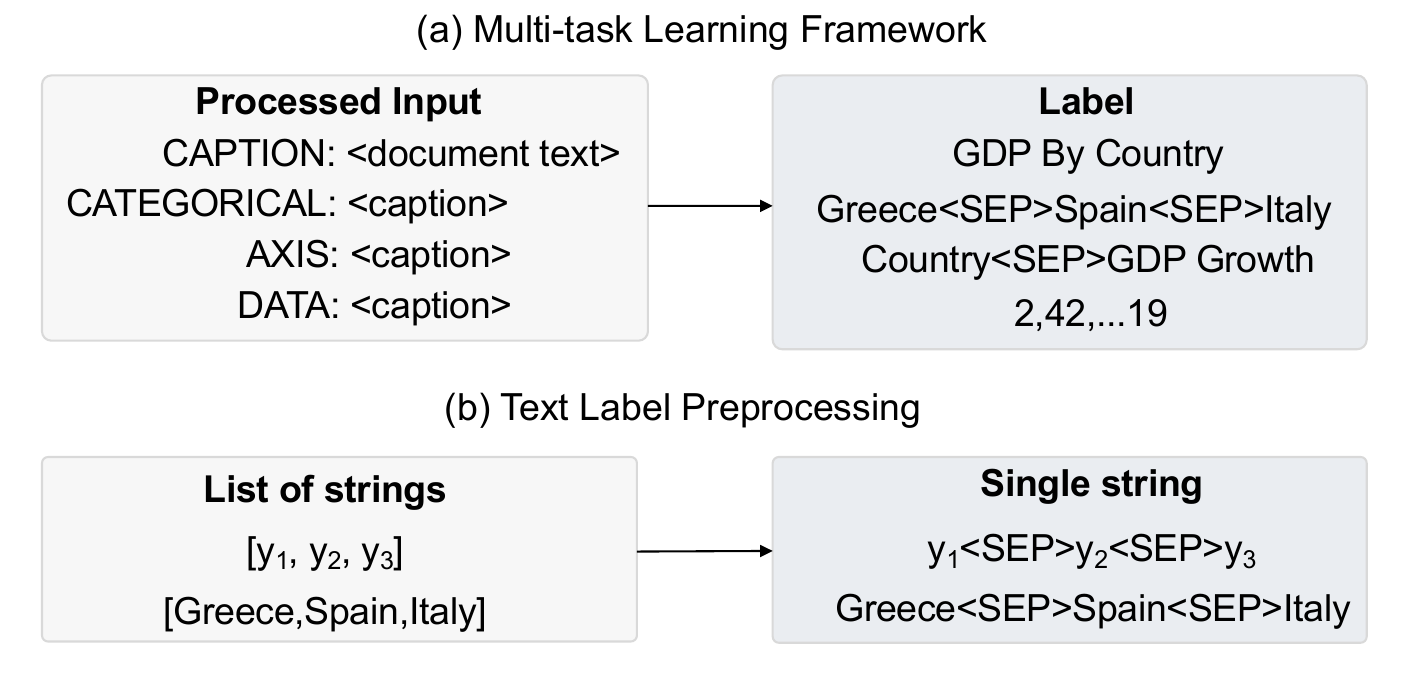}
  \caption{Preprocessing and multi-task learning framework.}
   \label{fig:multitasklang}
\end{figure}

\paragraph{Multitask Learning.}
The Transformer is simultaneously trained on tasks by pre-pending task-specific tags to the input to generate different outputs, as seen in Fig~\ref{fig:multitasklang}a. 
There are five different tasks: caption, axis, series names, categorical data and data tokens.
The first four tasks are output by the language decoder, while the continuous data task is output by the data decoder. 
This approach creates contextual consistency, improves performance due to the shared information between each task and saves computational resources.

 The training regime for the model is divided into two stages, one for each modality.
The first stage focuses on the language modality and trains the shared encoder and language decoder on the four language tasks. 
Only one of the four tasks is randomly selected for each instance during a training iteration.
The second stage focuses on the data modality and trains only the data decoder on the chart data task.
During this stage, the shared encoder and language decoder parameters are frozen, and only the data decoder is updated.

\subsubsection{\textbf{Pre-processing.}}

Random snippets of local document text are used as input to train the model to generate captions. 
The local text is extracted from the document based on the location of the figure. 
The nearest $N$ sentences from either side of the figure reference are fed into the model with probability $p$ of adding or removing sentences from each side. 
This constraint forces the model to generate realistic captions with varying amounts of context and reduces over-fitting.

The chart caption tends to have high semantic consistency with other text items in a chart. 
Semantic consistency means that the words and sentences address similar topics or related concepts and are thus close to one another in the language representation space \cite{mikolov2013distributed}. 
For example, if the axis title of a bar chart reads "Country", the reader would expect the caption to describe some statistics by country and the categorical data to contain a list of countries.

The caption is used as input to generate three types of text commonly found on charts: categorical labels, series names and axis titles.
The main challenge we encounter is that language models require the label to be formatted as a single natural language string.
Axis titles, series names and categorical labels are typically \emph{sequences of phrases or sequences of words}, therefore requiring specialized pre-processing to be converted into a single string.

To create this label, we concatenate each list of words appearing in the original chart training data using the special token <SEP> as a separator, as shown in Fig~\ref{fig:multitasklang}a.
This label is converted into a sequence of tokens using byte-pair encoding (BPE) \cite{gage1994new,sennrich2015neural}, which is then mapped to a sequence of embeddings using a look-up.
During training, the language decoder takes left-shifted embeddings as input and predicts the sequence of tokens using standard cross-entropy loss.
During generation, the string is converted back into the list of words using the special tokens as points of reference.

The Transformer also learns to map captions to chart data tokens and chart types (e.g. line or bar). 
To create a label, chart types are mapped to discrete indices and concatenated with the data tokens to form a single sequence of discrete indices. 
During inference, this sequence is sampled from the Transformer and passed to the \chartmodelname \, which generates the underlying chart data.

\section{Evaluation Setup}
\label{plots:sec:eval}
Our goal for the evaluation section is to determine the best components for the \honeyplotname \ by benchmarking them against reasonable baselines.
This section begins by describing the dataset in Section~\ref{sec:dataset}, the proposed baseline models in Section~\ref{sec:baseline} and then outlines the evaluation frameworks for each component: the Chart Data Model in Section~\ref{sec:expnonlang} and the multimodal multitask Transformer in Section~\ref{sec:explang}.


\subsection{Document-Chart Dataset}
\label{sec:dataset}

We introduce the first document-chart dataset to date, initially derived from the ICPR 2020 competition on harvesting raw tables \cite{davila2021icpr}.
This dataset contains the following chart features: axis names, series names, and continuous and categorical data. 
We harvested the associated caption and document text for each chart from PubMedCentral Open-Access to tailor this dataset to our specific requirements.
The figure captions were identified using the "figure" and "caption" tags found in associated XML files.
In total, there are 5,418 document-chart pairs containing raw data, captions, and document text.
These can be classified into five chart types, including line (1851), scatter (669), vertical bar (1666), horizontal bar (636) and box chart (596). 
These charts were then divided into an 80:20 train and test split.
The dataset can be downloaded from this link: \href{https://decoychart.s3.ap-southeast-2.amazonaws.com/document-chart-dataset.zip}{[Link]}. 

\subsection{Baseline Model}
\label{sec:baseline}
As there are no existing chart generation systems in current literature, we propose \textbf{ChatGPT} \cite{brown2020language} and its successor \textbf{GPT-4} \cite{gpt4techreport} as the baseline for both chart text and data generation due to their input flexibility.  
These Transformers were pre-trained on a large private dataset and fine-tuned using \emph{Reinforcement Learning from Human Feedback} \cite{ziegler2019fine}, enabling human-level conversation capabilities and performance on many challenging academic exams.

Our studies have found that both models can generate chart data using the document context as input. 
We use a carefully crafted prompt which allows the models to generate a structured response:

\begin{quotebox}
Based on the following text from a document, I need to create a \textless chart type\textgreater \ chart. Please provide me with a caption, series name, axis names, and data. I want everything in JSON format for Plotly. \\ Document Text: "\textless document text\textgreater " \\
If there are no specific values, then make them up based on the context. Feel free to make up your own values, too.
\end{quotebox}

The structured JSON format allows us to streamline the benchmarking process. 
Three different responses for each prompt were collected using the ChatGPT3.5 API \footnote{https://platform.openai.com/ (Accessed on 1--3 May 2023)}.
Multiple responses were required as the model can be unreliable, as discussed in the recent technical report by OpenAI \cite{gpt4techreport}. 
Following processing, 909 of 1083 responses had data correctly formatted for benchmarking.

We did not have API access to GPT4; therefore, we were limited to harvesting responses from the web browser \footnote{https://chat.openai.com/ (Accessed on 15--17 April 2023)}.
This made response capture slow and, as of writing, restricted users to only 25 responses per 3 hours.
For this model, we collected and cleaned 108 responses.


Note that since the contents of their massive internet-scrapped datasets remain private, it is unclear whether these models were trained on the test dataset. 
Despite this limitation, we find these models to be a very useful comparison where there are no alternatives.
Responses from baselines can be downloaded from \hyperlink{https://decoychart.s3.ap-southeast-2.amazonaws.com/gpt_responses/2023-05-01.zip}{[this link]}.

\subsection{Setup for Chart Data Model}
\label{sec:expnonlang}
There are no existing works that primarily focus on generating chart data.
For this reason, we benchmark several variants of our proposed Chart Data Model, each described in Section~\ref{sec:datamodel} and propose metrics in Section~\ref{sec:datametric}.
Our goal is to find a Chart Data Model from the list of candidates that generates samples with the highest reconstruction quality.

\subsubsection{Candidates}
\label{sec:datamodel}
\begin{enumerate} 
    \item \textbf{\honeyplotnameshort-VQVAE} combines VQVAE \cite{van2017neural} with the novel multi-head encoder and decoder. It uses a single latent codebook as the bottleneck.
    \item \textbf{\honeyplotnameshort-VQGAN} \cite{esser2021taming} introduces a generative adversarial network (GAN) and a min-max adversarial training objective. 
    This framework utilizes a discriminator, initialized as a multi-head encoder, which learns to distinguish between real and fake data by minimizing a binary cross-entropy loss function. 
\end{enumerate}
Each variant employs the same bottleneck size: a total codebook size of 256, a data token sequence length of 28 and a dimensionality of 32.


\subsubsection{Metrics}
\label{sec:datametric}

\paragraph{\textbf{Fréchet Inception Distance (FID)}} \cite{heusel2017gans}. 
FID was initially proposed to assess the realism of GAN-generated samples. 
In this study, we adapt FID to evaluate the similarity between authentic and fake continuous chart data. 
We employ a multi-head encoder, as described in Section~\ref{sec:encoder}, to distinguish between authentic and fake data. 
To produce fake chart data, we perturb existing chart data by randomly scaling each individual value using scales drawn from a uniform distribution, $\mathcal{X} \sim U(0.5, 1.5)$. 
The penultimate output layer consists of 32 units, and the FID distance between two samples is determined using the following calculation:
\begin{equation} \label{eq::fid}
    d^2 = ||mu_1 - mu_2||^2 + Tr(C_1 + C_2 - 2(C_1 \times C_2)^{1/2})
\end{equation}
where $mu_i$ is the vector mean, $C_i$ is the co-variance matrix, and Tr is the trace operation.
A smaller FID score is interpreted as better as it indicates higher similarity between the generated and authentic samples. 

The FID model's test accuracy at detecting fake chart data was 76\%. The code and weights have been open-sourced for others in the community to use for benchmarking.


\subsection{Setup for Multimodal Transformer Model}
\label{sec:explang}
We benchmark the performance of three multimodal Transformer models designed according to the framework in Section~\ref{sec:charttextmodel}.
The base pre-trained Transformer models are described in Section~\ref{sec:langmodels}, followed by two metrics in Section~\ref{sec:langmetrics}.
The goal is to find a multimodal Transformer model from the list of candidates that can generate both realistic chart text and data.


\subsubsection{Candidates} 

\label{sec:langmodels}
\begin{enumerate}
    \item \textbf{\honeyplotnameshort-T5} is based on Text-to-Text Transfer Transformer (T5) \cite{raffel2020exploring}, a highly scalable sequence-to-sequence Transformer pre-trained on a large corpus harvested from Common Crawl. 
    T5 introduced a novel pre-training objective that randomly corrupted spans of text using masks.
    Raffel et al. demonstrate that T5 outperforms prior language models across multiple benchmarks, such as GLUE and SquAD. 
    We employ the updated T5 v1.1 large model from HuggingFace \cite{wolf2019huggingface}, which introduces GEGLU activation in feed-forward hidden layers.
    Our version with two decoders contains 919M total parameters.
    \item \textbf{\honeyplotnameshort-Pegasus} \cite{zhang2020pegasus} is a sequence-to-sequence Transformer developed for text summarization. 
    Pegasus was pre-trained by randomly masking whole sentences (gap sentences) and then fine-tuned on 12 different datasets, demonstrating state-of-the-art performance across multiple ROUGE metrics. 
    As our dataset is derived from PubMedCentral, we modified the version fine-tuned on the same corpus, which consists of a total of 837M parameters.     
    \item \textbf{\honeyplotnameshort-Big Bird} \cite{zaheer2020big} is another sequence-to-sequence Transformer that uses sparse block attention to address longer sequences of text.  
    Sparse block attention improves the original attention mechanism \cite{vaswani2017attention} by reducing the memory dependency on sequence length from quadratic to linear. 
    The authors demonstrate improved performance over Pegasus across multiple tasks, including summarization and question answering.
    We modified the version of Big Bird fine-tuned on PubMedCentral with 849M parameters in total.
    \end{enumerate}
Language models are implemented using the HuggingFace library, Pytorch \cite{paszke2019pytorch} and were trained across all tasks.
The models are trained on the language and data stages for 50 epochs each using the AdamW optimizer \cite{loshchilov2017decoupled}, a learning rate of 5e-4, cosine annealing learning schedule and batch size of 64.
The document text was extracted as 16 sentences from both sides of the first mention of the chart.
During training, the sentence count was randomly reduced or lengthened by one at a fixed probability of 10\%.

\subsubsection{Metrics}
\label{sec:langmetrics}
\begin{enumerate} 
    \item \textbf{Recall-Oriented Understudy for Gisting Evaluation (ROUGE).} \cite{lin2004rouge} This metric is widely used to benchmark the natural language models on the challenging problem of text summarization \cite{zaheer2020big,zhang2020pegasus,raffel2020exploring}. 
    We propose ROUGE because figure captions can typically \textit{summarize} the key points of the surrounding document text.
    ROUGE-N measures the overlap of N-grams between the reference and generated text. 
    ROUGE-L refers to the "longest common sub-sequence", the longest sequence of words appearing in the same order in both the reference and generated text.
    A higher ROUGE score is considered better. 
    
    However, we do not consider this metric sufficient on its own, therefore, we propose a second metric called Keyword Semantic Matching, which measures semantic similarity using embedding spaces.
    \item \textbf{Keyword Semantic Matching (KSM).} This captures the semantic similarity between a list of reference topics (or keywords) and generated text. This novel metric is a variant of Topic Semantic Matching (TSM) \cite{timmer2022tsm}, which measures semantic similarity between real documents and honeyfiles.
    KSM extracts topic representations using YAKE \cite{campos2020yake} and employs a different embedding method as described below. 
    We use KSM to compare honey file text to generated caption text and compare caption text to chart text elements.
    KSM has two inputs, a reference text and a generated text, and returns a measure of semantic distance between the texts.
    \begin{enumerate}
        \item Keywords are extracted from the texts using YAKE. If either reference or generated text is short (typically under 20 words), keyword extraction is skipped, and the whole text is treated as a single keyword. \footnote{We refer both single keywords and phrases or sequences of words as \emph{keywords} because they are eventually treated as multiple tokens in language models, so there is no practical difference.}
            \item Both reference and generated keyword lists are tokenised with BPE and converted into an embedded representation by encoding with a pre-trained language model encoder. We use Pegasus because it was fine-tuned on PubMedCentral, the same source as our dataset.
        \item Embeddings for each keyword are averaged, leaving a single latent vector.
        This approach provides a better latent representation in comparison to only averaging over the sequence of embeddings \cite{ni2021sentence}.  
    
        \item We compute the cosine distance between each pair of references and generate keyword embeddings.  The cosine distances are averaged into a semantic distance between the reference and generated texts.  A higher number indicates greater similarity. 
    \end{enumerate}
   
\end{enumerate}

\section{Results}
\label{plots:sec:results}

In this section, we present the performance of the \honeyplotname \ according to the proposed evaluation setup.
We conduct two experiments for the Chart Data Model and Multimodal Transformer against baselines and present results and analysis for each.

\subsection{Chart Data Reconstruction}
\begin{table}[h]
\centering
    \caption{FID scores for chart data generation.}
  \begin{tabular}{lcc}
    \toprule
     &Train FID&Test FID\\
     \midrule
     \honeyplotnameshort-VQVAE&17.16&13.45\\
     \honeyplotnameshort-VQGAN&14.55&11.58\\
    \bottomrule
    \end{tabular}
  \label{tab:data}
\end{table}

\begin{figure*}[h]
  \centering
  \includegraphics[width=\linewidth]{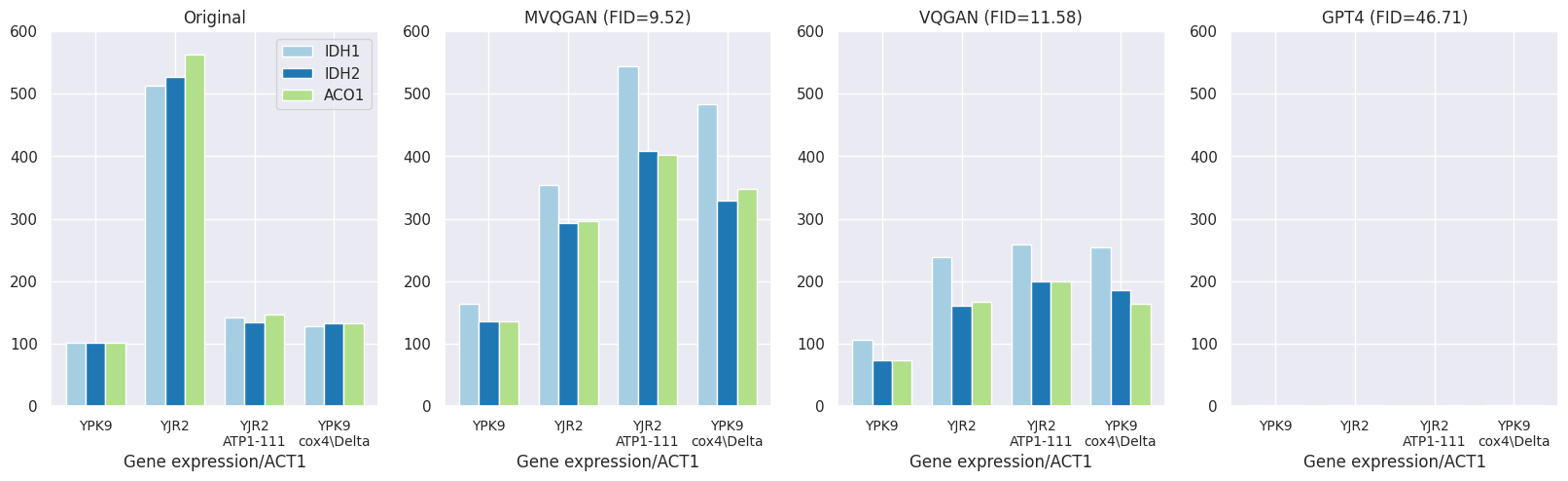}
  \caption{A comparison of original chart data to model reconstruction. GPT4 is poor at determining the correct data range, which leads to low FID scores.}
   \label{fig:recons1}
\end{figure*}

\begin{table*}[h]
\small
\centering
  \caption{ROUGE-1/ROUGE-2/ROUGE-L scores across each language task. }  
  \begin{tabular}{lccccc}
    \toprule
     Model & Caption & Series Name & Axis Title & Categorical \\
     \midrule
     ChatGPT & 0.192/0.069/0.168 & 0.065/0.030/0.058 & 0.150/0.056/0.143  & 0.114/0.041/0.079 \\ 
     GPT4 & 0.220/0.083/0.193 & 0.106/0.032/0.107 & 0.162/0.056/0.160   & 0.141/0.030/0.103 \\
    \honeyplotnameshort-Pegasus& 0.213/0.085/0.179 & 0.019/0.001 /0.019  & 0.076/0.019/0.072 & 0.081/0.018/0.064 \\
    \honeyplotnameshort-Big Bird & 0.235/0.084/0.199 & 0.125/0.052/0.115  & 0.254/0.100/0.244  & 0.115/0.057/0.108 \\
    \honeyplotnameshort-T5 & \textbf{0.294/0.137/0.256} & \textbf{0.281/0.166/0.249} & \textbf{0.378/0.186/0.354} & \textbf{0.198/0.102/0.191} \\
    \bottomrule
    \end{tabular}
  \label{tab:rougescores}
\end{table*}

\begin{table*}[t]
\small
\centering
  \caption{Keyword Semantic Matching and Fréchet Inception Distance (FID) scores for each task. }  
  \begin{tabular}{lccccccc}
    \toprule
      & \multicolumn{5}{c}{Keyword Semantic Matching} & \multicolumn{2}{c}{FID} \\
     \cmidrule(r){2-6}
     \cmidrule(r){7-8}
     Model & Caption & Series Name & Axis Title & Categorical & Average & Train & Test  \\
     \cmidrule(r){1-6}
     \cmidrule(r){7-8}
     ChatGPT & 0.725 &0.835 &0.792 &0.826& 0.794 &39.91&34.47\\ 
     GPT4 &0.698 & 0.774 & 0.845 &0.832 & 0.787&55.83&46.71\\
    \honeyplotnameshort-Pegasus&0.850 & 0.758 & 0.802 & 0.800 & 0.802 & 44.14 & 41.29\\
    \honeyplotnameshort-Big Bird & \textbf{0.858} & 0.820 & 0.835 & 0.821 & 0.833 & 40.44 & 34.96\\
    \honeyplotnameshort-T5 & 0.851 & \textbf{0.836} & \textbf{0.850} & \textbf{0.848} & \textbf{0.846} &  \textbf{37.62} & \textbf{32.14} \\
    \bottomrule
    \end{tabular}
  \label{tab:ksmscores}
\end{table*}

In our first experiment, we evaluate each chart data model's ability to reconstruct chart data using FID.
The results are shown in Table~\ref{tab:data}, demonstrating that the best-performing candidate was VQGAN, with a test FID score of 11.58.

All models perform better on test FID because the inputs were sampled from the test dataset.
In Fig~\ref{fig:recons1}, a comparison of charts is presented after rendering with matplotlib.
These figures demonstrate that the ability to predict the scale significantly impacts the FID scores, as seen by comparing the y-axis.

As the \honeyplotnameshort \ variants separate scale parameters from the continuous data, the outputs are more realistic in comparison to purely text-based models.
This is particularly important because the range of values must stay within an acceptable range according to the honeyfile context.
If the chart values stray outside these ranges, it may raise suspicion of intruders.


\begin{table*}[t]
\small
    \caption{Comparison of generated captions using the same input by various Transformer language models. The document text was cropped to fit in page limits.}
  \begin{tabular}{p{2cm}p{10cm}}
    \toprule
     \multicolumn{2}{l}{\bfseries Document Text (Input) } \\
     \midrule
    \multicolumn{2}{p{12cm}}{The mean and median ages of the study subjects were 64.7 and 55 months, respectively. ... No measles-specific IgG antibody was detected among infants aged 3-8 months. 
    Although three infants in this age group (one aged 6 months and two aged 8 months) had documented proof of measles vaccination, none of these infants had detectable measles-specific IgG antibodies (Figure 1). Measles-specific antibody response among children \textgreater9 months of age.
    The presence of measles-specific IgG antibodies increased with age to 57.3\% (43/75) ... 
    } \\
    \midrule 
    \bfseries Model& \bfseries Caption (Output) \\
    \midrule
    Label&Presence of measles-specific IgG antibody, infants aged 0-8 months \\ \midrule
    \honeyplotnameshort-T5&Measles-specific antibody response among children \textgreater9 months of age.\\  \midrule
    \honeyplotnameshort-Big Bird&The antibody response in children aged 6 months to 5 years, by immunization status. \\  \midrule
    ChatGPT&Vaccination Status by Age.\\  \midrule
    GPT4&Vaccination Card Possession by Age Group.\\  \midrule
    \bottomrule
    \end{tabular}
  \label{fig:captionoutputs}
\end{table*}
\begin{table*}[t]
\small
    \caption{Comparison of generated chart text by various language models. GPT-based models tend to perform better at generating categorical data but less so at other tasks.}
  \begin{tabular}{p{1.6cm}p{4.5cm}p{2.7cm}p{3.0cm}}
    \toprule
     \bfseries Model &  \bfseries Categorical Data & \bfseries Series Name & \bfseries Axis Title   \\
     \midrule
    Label & 0-8 months, 9 months-5 years, 6-9 years, \textgreater10 years & Presence of IgG\newline Absence of IgG & Age group \newline Percentage (\%) \\ 
    \midrule
    \honeyplotnameshort-T5 & 0 1 2 3 4 5 6 7 8 9 10 11 12 \newline 13 14 15 16 17 18 19 20  & 0-9 months \newline
    9-10 months \newline
    11-13 months \newline
   14-16 months
   & Antibody response\newline Age (months)\\ 
    \midrule
    \honeyplotnameshort-Big Bird&6 months to 5 years, 1 year, \newline 2 years, 3 years, 4 years, 5 years
    & Age (g/m2) \newline Age (months)&Time (years)\newline Relative antibody titer\\ 
    \midrule
    ChatGPT&Reported Routine Coverage: 58.2\%\newline Routine or Catch-up Campaign Coverage: 83.1\% & Routine Coverage & Age (months) \newline Vaccination Coverage \newline  (\%)\\ 
    \midrule
    GPT4& \textless 1 year, 1-2 years, 3-8 months,\newline 9 months-5 years, \textgreater10 years& Number of children \newline with vaccination cards &Age Group \newline Number of Children\\ 
    \bottomrule
    \end{tabular}
  \label{fig:textoutputs}
\end{table*}

 \subsection{Multimodal Transformer Performance}
In this experiment, the proposed multimodal Transformer models were jointly trained on the four language tasks and data tokens sampled from \honeyplotnameshort-VQGAN.
Table~\ref{tab:rougescores} presents the ROUGE scores across each model and task.
In summary, the \honeyplotnameshort-T5 achieved the highest average ROUGE score of 0.232, outperforming both baselines: GPT4 (0.116) and ChatGPT (0.097).
\honeyplotnameshort-T5 also demonstrates the highest ROUGE scores across all individual language tasks.
\honeyplotnameshort-Pegasus achieved the lowest average score of 0.070; however, it is a relatively competitive score for the caption generation task compared to baselines.
This is likely because Pegasus was designed specifically for summarization, a similar problem.

The KSM and FID scores for each model are also presented in Table~\ref{tab:ksmscores}. 
These results show that \honeyplotnameshort-T5 also performs the overall best with an average KSM score of 0.846 and an FID score of 32.14.
It also achieves the highest score across all individual tasks except caption.
\honeyplotnameshort-Big Bird marginally outperforms \honeyplotnameshort-T5 on the caption task due to its use of sparse block attention.
Recall that sparse block attention allows Big Bird to address longer spans of text and that KSM measures the semantic similarity between the local document text and generated caption.
For other language tasks, KSM measures the similarity between the caption and the generated list of text. 
Captions are vastly shorter than document text, which reduces the advantage of sparse block attention for these tasks.

The performance difference between the three candidate Transformers could be attributable to how they were pre-trained.
Pegasus and Big Bird were trained for abstract summarization, while T5 was trained for a wider variety of tasks, which likely contributed to its adaptability when learning new tasks and modalities.
The \honeyplotnameshort-T5 also has more parameters overall (919M), which gave it more capacity to fit the new data token distribution.

In Table~\ref{fig:captionoutputs}-\ref{fig:textoutputs}, we provide text snippets from each model using the same local document text as input.
All models can generate semantic consistent captions that identify the two main topics: infants and antibodies/vaccinations.
The \honeyplotnameshort \ Transformers produce more specific captions in comparison to the GPT models. 
This can be seen quantitatively in Table~\ref{tab:rougescores}-\ref{tab:ksmscores}, where ChatGPT and GPT4 achieve slightly lower scores across both ROUGE and KSM. 

We observe that \honeyplotnameshort \ Transformers can be prone to generating similar text for the categorical, series name and axis title tasks. 
This can be seen in Table~4 where the axis title and series name both mention "months".
This problem doesn't appear in ChatGPT and GPT4 outputs, as all chart text is generated in a single sequence. 
A possible solution for this would be to train the \honeyplotnameshort \ Transformer by conditionally generating each task in a multi-task forward chain.
The chain should start with larger concepts, such as axis titles or series names and then end with categorical or continuous data.
We leave this as a future direction of research.

In this work, we considered a subset of all possible chart parameters, which may also include the colours, text size and font types.
It is possible to include these additional parameters as part of the multi-task framework discussed above; however, it can be challenging to harvest this data.
Pixel-based models automatically learn these parameters however are poor at generating text as seen in Fig~\ref{fig:pixelcomparison} for both Stable Diffusion \cite{rombach2022high} and DALLE2 \cite{ramesh2022hierarchical} \footnote{Accessed from stablediffusionweb.com, labs.openai.com and chat.openai.com on 21 May 2023.}.

\section{Discussion and Challenges}

Our investigation into honeyplot automation explored the combination of various generative modelling approaches to achieve our main objective of document realism.
Here, we summarize our most important findings and discuss challenges that may offer guidance to others interested in researching honeyfiles.

Our most important finding is that large language models are adept at generating coherent and semantically consistent chart text and can also connect to discrete autoencoders to generate chart data. 
This capability is possible through a shared encoder, a secondary decoder and a two-stage training regime.
This finding allowed us to address the key limitations identified in Sec~\ref{sec:intro}. 

Our second finding is that multiple charts with different data structures can be simultaneously trained using the same discrete autoencoder.
This was accomplished by employing multiple heads to separately generate the data scales and data structure shape.
Our evaluations demonstrate that the multi-head design can learn the chart data scales and patterns significantly better than large language models.

An important consideration for using downstream visualization software is that the defending entities, such as corporations and governments, commonly employ recommended document-style templates.
Pixel-wise generative models could theoretically generate visual aspects of the honey chart using style transfer methods.
However, this approach would require ample amounts of chart data specific to the defending entity for training, which may be difficult to collect or unavailable in sufficient quantity.
Tailoring the visualization software in HoneyPlotNet's pipeline to employ these style templates would improve the deceptiveness of the honeyfiles as they would blend into the population of existing documents.

A major challenge we faced was designing a large-scale user study to support a robust qualitative evaluation for our experiments. 
There were many obstacles, such as setting up adequate deception environments and recruiting sufficient users that accurately reflected the realistic target of our work: hackers.
These challenges have also been documented in the work of Ferguson et al. \cite{ferguson2018tularosa}.
To address this concern, we will be open-sourcing the dataset,  \honeyplotnameshort \ codebase and will be supporting anyone interested in a large-scale capture-the-flag exercise. 

\section{Conclusion}
\label{plots:sec:conclusion}
This paper introduced  \honeyplotname \, a generative model that produces deceptive charts (honeyplots) based on the local document context.
This contribution is very useful for defenders aiming to improve the automation and realisticness of their honeyfiles.
Our approach combines a discrete autoencoder and a large language model with a dual-decoder, which results in better performance than single-decoder language models such as ChatGPT. 
The release of the first-ever document-chart dataset represents a valuable resource for the community. 
Additionally, our proposed Keyword Semantic Matching (KSM) metric fills a critical gap in honeyfile evaluation.
We believe these contributions not only advance the field of honeyfile generation but also pave the way for further research and development in defensive cybersecurity strategies.






\section{Acknowledgements}
The authors would like to acknowledge the support of the Commonwealth of Australia and the Cybersecurity Cooperative Research Centre.

\bibliographystyle{ACM-Reference-Format}
\bibliography{references}


\end{document}